\documentclass{article}
\usepackage{arxiv}
\usepackage{floatrow}
\newfloatcommand{capbtabbox}{table}[][\FBwidth]
\usepackage[utf8]{inputenc} 
\usepackage[T1]{fontenc}    
\usepackage{hyperref}       
\usepackage{url}            
\usepackage{booktabs}       
\usepackage{amsfonts}       
\usepackage{nicefrac}       
\usepackage{microtype}      
\usepackage{lipsum}		
\usepackage{graphicx}
\usepackage{natbib}
\usepackage{doi}

\usepackage{times}
\usepackage{soul}
\usepackage{url}
\usepackage{etoolbox}
\usepackage{booktabs}
\usepackage{multirow}
\usepackage{algorithm}
\usepackage{algpseudocode}
\usepackage{amsmath}
\usepackage{amssymb}
\usepackage{latexsym}
\usepackage{graphicx}
\usepackage{longtable}
\usepackage{tabularx}
\usepackage{diagbox} 
\usepackage{todonotes}
\usepackage{caption}
\title{Causal Feature Selection with Dimension Reduction\\ for Interpretable Text Classification}

\author{
Guohou Shan$^1$
James Foulds$^{2}$
Shimei Pan$^2$\footnote{Contact Author}\\
$^1$Temple University\\
$^2$University of Maryland, Baltimore County\\
\ tul05519@temple.edu,
\{jfoulds,shimei\}@umbc.edu
}

\renewcommand{\shorttitle}{Causal Feature Selection}

\hypersetup{
pdftitle={A template for the arxiv style},
pdfsubject={q-bio.NC, q-bio.QM},
pdfauthor={David S.~Hippocampus, Elias D.~Striatum},
pdfkeywords={First keyword, Second keyword, More},
}

\begin{document}
\maketitle

\begin{abstract}
Text features that are correlated with class labels, but do not directly cause them, are sometimes useful for prediction, but they may not be insightful. 
As an alternative to traditional correlation-based feature selection, causal inference could reveal more principled, meaningful relationships between text features and labels. 
To help researchers gain insight into text data, e.g. for social science applications, in this paper 
we investigate a class of matching-based causal inference methods for text feature selection. Features used in document classification are often high dimensional, however existing causal feature selection methods use Propensity Score Matching (PSM) which is known to be less effective in high-dimensional spaces.  
We propose a new causal feature selection framework that combines dimension reduction with causal inference to improve text feature selection. Experiments on both synthetic and real-world data 
demonstrate the promise of our methods in improving classification and enhancing interpretability. 
\end{abstract}

\section{Introduction}
In text classification, representing text often requires a large number of features. For example, TF*IDF, and $n$-grams are frequently used in  representing documents. Since even the number of unigram or TF*IDF features is equal to the number of unique words in a text corpus, we can easily extract thousands or even millions of features from text. 
Moreover, when the ground truth data needed to train a text classifier is limited, the classifier can easily overfit the training data and perform poorly on new data.  Dimension reduction methods are thus important for improving both accuracy and interpretability. 

Feature selection is a common approach for dimension reduction in which noisy features that may confuse a classifier are removed before or during training. Compared with projection-based dimension reduction methods such as PCA ~\cite{li2014muli} or neural network-based embedding methods such as BERT ~\cite{devlin-etal-2019-bert} which learn new latent features that may be hard to interpret, 
feature selection identifies text features that are directly interpretable, since word-based features are meaningful to humans by definition.    
The most commonly used feature selection techniques are correlation-based~\cite{hall1999correlation,yu2003feature}. Since an observed correlation could be due to uncontrolled confounding variables, they are not ideal for providing insight. For example, an observed positive correlation between \textit{ice cream sales} and \textit{shark attacks} can be misleading as this may be caused by a confounding variable, \textit{temperature increase}. This can also happen with text, e.g. \textit{ice cream sales}, \textit{shark attacks}, and \textit{temperature increase} are now $n$-grams.  Since causal inference can remove spurious associations 
and uncover principled relationships between input features and output labels, it has the potential to provide better insight than correlation-based feature selection.

The only prior work known to us on causal text feature selection is that of ~\cite{paul2017feature}, who uses Propensity Score Matching (PSM) for causal inference~\citep{Rubin1974}.  
However, PSM is not ideal for high-dimensional data such as text~\cite{King2019}. In this study, we propose a novel approach to address the main limitations of~\cite{paul2017feature}.  The main contributions of our research are:
\begin{itemize}
    \item We develop and formalize a causal feature selection framework which generalizes, makes precise, and specifies the assumptions made by an approach previously considered by~\cite{paul2017feature}. 
    
    \item We develop a new causal text feature selection method, including a principled approach to leverage dimensionality reduction in the process, which addresses one of the main limitations in~\cite{paul2017feature}.  
    %
    
    
    \item  We evaluate our methods on two real-world 
    and two synthetic datasets. Our results demonstrate that our methods achieve better classification performance as well as improved interpretability compared to widely used  correlation and causal-inference based methods.
\end{itemize} 
 
\section{Related Work}
Most existing work on text feature selection is correlation-based~\cite{le-ho-2015-comprehensive,agarwal-mittal-2012-categorical,hall1999correlation}. The only work known to us on causal text feature selection employs Propensity Score Matching (PSM)~\cite{paul2017feature}. Since PSM was shown to not perform well in high dimensional spaces due to issues such as bias, statistical inefficiency, and imbalanced matching ~\cite{Gu1993,Caliendo2008,King2019}, it is not ideal for text feature selection. 

In NLP, there is also some work on extracting causal relationships expressed in text~\cite{dasgupta-etal-2018-automatic-extraction,mirza-tonelli-2016-catena}. However, our
task is fundamentally different. Rather than identifying causal relations expressed in text, which is an Information Extraction (IE) task, we are using causal models to identify significant features that have causal relations with outcome variables.

There is also work on applying NLP techniques to aid causal inference~\cite{wood-doughty-etal-2018-challenges}, and on applying causal inference to aid other NLP tasks~\cite{Landeiro2016,roberts2018adjusting}. For example, ~\cite{roberts2018adjusting} proposed a matching method called Topical Inverse Regression Matching for causal inference controlling for text data.  Their work does not study feature selection. 

\section{Background: Causal Inference}
In this section, we describe the causal inference framework our research is based on. We also describe two widely used causal inference methods instantiated from this framework. 
 
\subsection{Neyman-Rubin Potential Outcomes Framework}
The framework is a popular causal inference formulation
~\cite{Neyman1923,Rubin1974}. Within the framework, for a sample $i$,  there are two possible outcomes (e.g., class labels): $Y_{i}(1)$ if $i$ undergoes treatment $T$ and $Y_{i}(0)$ if not.
The \emph{treatment effect} for observation $i$ is defined by $Y_{i}(1)$-$Y_{i}(0)$. Since for each sample $i$, we observe either $Y_{i}(1)$ or $Y_{i}(0)$, one of the two conditions is always missing in observational data. 

A naive approach for testing the effect of a treatment $T$ is to directly compare the treated and control groups. This works in randomized experiments where the treatment
assignment is random. Since random assignment of samples to treatment groups balances both known and unknown sample characteristics that may affect the outcome, it reduces the likelihood that there will be differences in the sample characteristics between treated and control groups. In an observational study, however, this balance can be distorted by 
the systematic assignment of the treatment to samples~\cite{rosenbaum1983central,pearl2010,Hill2013}. Thus, the \emph{average treatment effect (ATE)} for
the treated group cannot be directly estimated.  
To address this issue, typical approaches aim to counteract the relationship between treatment and covariates by matching instances in the treatment and control groups. 

\subsection{ Mahalanobis Distance Matching (MDM) and Propensity Score Matching (PSM)}

The most straightforward and nonparametric way to match treatment and control instances is to exactly match on the pre-treatment covariates $X$.  
This approach is infeasible if the sample size is limited and the dimensionality of $X$ is large or if $X$ contains continuous covariates.  
Coarsened exact matching (CEM) performs binning on each variable before matching identical instances~\cite{Iacus2012}, but does not address the curse of dimensionality.  
Two common approaches to solve this problem are nearest neighbor matching (NNM), often based on the Mahalanobis distance (MDM) to account for scaling and correlations~\cite{cochran1973controlling,Rubin1979}, and propensity score matching (PSM)~\cite{rosenbaum1983central}. 
The Mahalanobis distance between any two samples $X_i,X_j$ is defined as $$MD(X_i,X_j)=\{(X_i-X_j)^{'}S^{-1}(X_i-X_j)\}^{1/2}$$ where $S$ is the empirical covariance matrix of $X$. In MDM, to estimate the treatment effect, one matches each treated sample with the closest control sample, as calculated by the Mahalanobis distance.

Another widely used matching approach is Propensity Score Matching (PSM)~\cite{rosenbaum1983central}. PSM matches on the probability of assignment to
treatment, known as the propensity score: 
$e(X_i)=Pr(T_i=1|X_i) \mbox{ .}$
It is typically estimated using logistic regression. 
The unidimensional metric acts as a one-number summary of the covariates, such that conditioning on the propensity score $e(X_i)$ leads to conditional independence between the features in the treatment and control groups, thereby approximating a randomized controlled experiment. 
Although very popular, PSM has some limitations.
For example, 
~\cite{king2016propensity} argue that PSM leads to more imbalance between treatment and control groups than matching methods that approximate full blocking such as MDM and CEM.  PSM also exhibits poor performance in high-dimensional spaces due to biased estimation, imbalance, and statistical inefficiency~\cite{Gu1993,Abadie2006,Caliendo2008,King2019}. 

In the next section, we propose a causal feature selection framework for text classification, which generalizes and formalizes the method proposed by~\cite{paul2017feature} and aims to address some of its limitations. 

\section{Problem Formulation}
\begin{figure}[t]
    \centering
    \includegraphics[width=15cm]{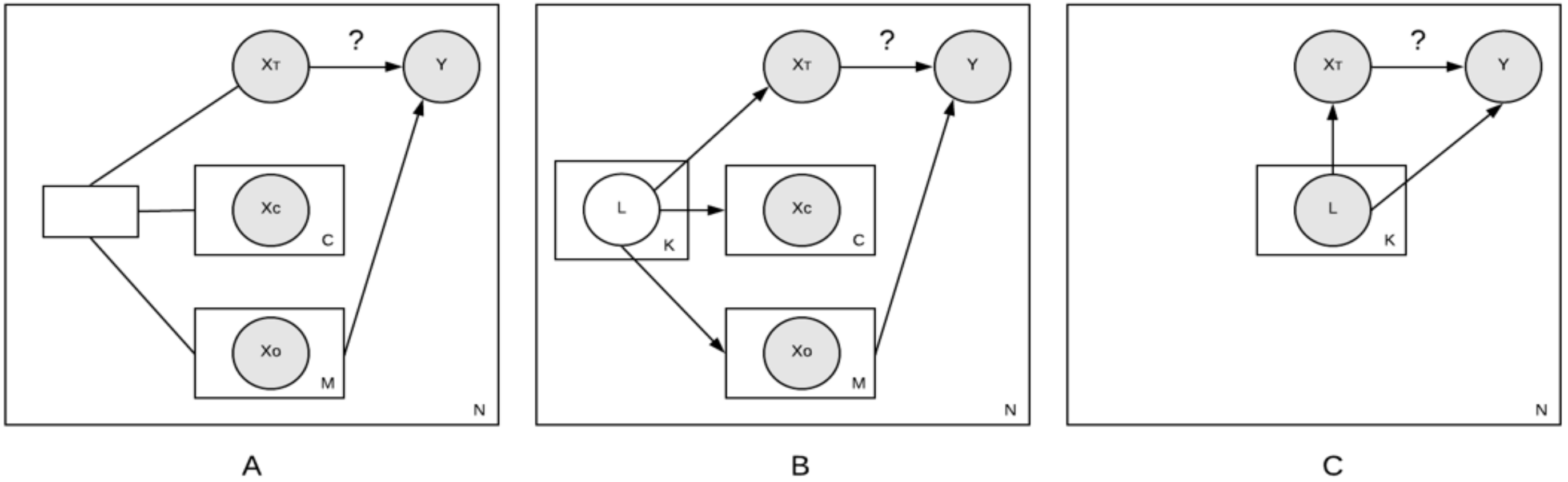}
    \caption{Assumptions of our Causal Feature Selection (CFS) framework. (A) Causal feature selection with observational features. (B) Causal feature selection with latent variable assumption. (C) Causal feature selection with dimensionality reduction. }
    \label{causalprocess}
\end{figure}
Figure \ref{causalprocess} illustrates our problem formulation for causal feature selection in three scenarios.   In the general case (Figure~\ref{causalprocess} \emph{(A)}), we assume that there are classification features $X$ (e.g., bag-of-words or TF*IDF features for a document), and an outcome $Y$ (i.e., the class label).  The features are divided into two main types: causal features $X_C$, which directly cause the outcome $Y$, and other non-causal observed features $X_O$ which have no direct causal relationship with $Y$.  We select one of the features as a ``treatment'' variable $X_T$, assumed to be binary, e.g. whether a word is present in a document.  Our goal is to use causal inference to determine whether the treatment feature $X_T$ directly causes the outcome $Y$, i.e. is $X_T$ a member of $X_C$, or of $X_O$?  In the full causal feature selection process, each feature will be considered as a ``treatment'' in turn.  The features that are found to cause $Y$ will be selected by our method. 

Figure~\ref{causalprocess} \emph{(A)} shows a graphical model diagram that describes the assumptions we make about the data generating process when applying our causal feature selection method.  The diagram combines factor graph notation, where a rectangular factor node indicates a clique of dependent variables, with directed edges that represent causal relationships.  We assume that $X_O$, $X_C$, and $X_T$ are related to each other in unknown ways, as indicated by their shared factor node in the factor graph. The directed edge from $X_C$ to $Y$ indicates that $X_C$ has a direct causal effect on $Y$.  An edge with a question mark is shown from $X_T$ to $Y$ since $X_T$ may potentially directly cause $Y$, and our goal is to verify whether this is the case.  All the variables in the graph are observed in this setting, so their nodes are shaded.

In a perfect world, we would have ``twin'' instances that differ only in $X_T$ so that we could observe both potential outcomes $Y_i(1)$ and $Y_i(0)$ for each instance $i$ to directly estimate the average treatment effect (ATE), and hence verify whether $X_T$ is causal for $Y$.  In practice, we can only observe one outcome per instance, and the treatment is also correlated with the other variables, unlike in a randomized controlled trial.  We must correct for this using causal inference. 

Previously, \cite{paul2017feature} proposed a particular instantiation of our general framework in Figure~\ref{causalprocess} \emph{(A)} in which \emph{PSM is applied to word-level features $X$.}

On the other hand, it is well known that causal inference techniques such as PSM struggle with data that are high-dimensional~\cite{Gu1993,Abadie2006,Caliendo2008,King2019}, as is common with word-level text features.  Instead, we propose to leverage a latent variable model (LVM) to reduce the effective dimensionality of the causal inference problem by inferring the latent variables and using them as the features for causal inference algorithms,
thereby avoiding the problems with PSM and related techniques in high-dimensional text data.

We aim to simplify the complex high-dimensional causal inference problem in Figure~\ref{causalprocess} \emph{(A)}.  To this end, we can consider simplifying assumptions on the joint distribution of $X$.  
A common approach used by latent variable models
is to assume that the relationships between the features $X$ are mediated by latent variables $L$, which provide low-dimensional representations of high-dimensional data such as text. For example, in the LDA topic model the complex dependencies between all words in a document are simplified by assuming that they are generated independently given a latent distribution over topics.
Figure~\ref{causalprocess} \emph{(B)} shows a latent variable model instantiation of Figure~\ref{causalprocess} \emph{(A)}, where there are $K$ latent variables $L$, e.g. the document's proportions for $K$ topics, which explain the features $X$. 
After assuming an LVM model as in Figure~\ref{causalprocess} \emph{(B)}, we arrive at our formulation for causal feature selection method with dimension reduction by marginalizing out $X_O$ and $X_C$, then representing the instances via the latent variables $L$ instead of the observable features (Figure~\ref{causalprocess} \emph{(C)}).  To determine whether $X_T$ is causal of $Y$, we now only need to control for differences in the lower-dimensional latent features using causal inference techniques, which is expected to be typically easier to achieve compared to using high-dimensional $X$.  To implement this in practice, we will use an algorithm to infer, hence effectively ``observe'' $L$ based on $X$.

Note that causal inference controlling for $L$ is different to causal inference controlling for $X$, and may lead to different conclusions.  However, since we assumed that $X$ was generated based on $L$, in many cases it is reasonable to assume that $L$ captures much of the important information in $X$.  Furthermore, for many NLP models $L$ is semantically meaningful.  For example, in LDA $L$ represents the topical content of the document, and conditioning on this representation may just as desirable, or even more desirable, compared to directly controlling for the differences in the frequencies of other word count features $X$. 
\section{Proposed Methods}
In this section, we introduce our proposed causal feature selection (CFS) method which includes four key steps: dimension reduction, matching, feature selection and classification. 

\subsection{Dimension Reduction}
Given all of the observed features $X$ and the latent dimensionality $K$, we employ dimension reduction to uncover $L$ which captures the main characteristics of each sample $i$.  
Our hypothesis is that 
matching based on $L$ will help us find better treated and control pairs than directly matching on surface features $X$, especially when $|X|$ is large. We explore a diverse set of dimension reduction methods including Principal Component Analysis (PCA), Gaussian Random Projection (GRP) and Mini-Batch Dictionary Learning (MBDL). 
%
Specifically,  \textbf{PCA} learns a low-dimensional representation of $X$ under independent identical Gaussian noise. It is one of the most widely used methods for dimension reduction~\cite{zhang2019matrix}. The normal version of PCA (\textbf{nPCA}) learns a linear combination of $X$ while sparse PCA (\textbf{sPCA}) uses Lasso (or Elastic Net) to derive modified principal components with sparse loading~\cite{zou2006sparse}. Moreover, random projection is a powerful method for dimension reduction, which theoretically preserves distances in the original feature space~\cite{bingham2001random}. When the random projection is produced by Gaussian distributions, it becomes Gaussian Random Projection (\textbf{GRP}). Finally, \textbf{MBDL} learns a dictionary (a set of atoms) that can be used to represent data using a sparse code~\cite{mairal2009online}. In addition, for text feature selection, we also employ  dimension reduction techniques widely used in text analysis, Latent Dirichlet Allocation (\textbf{LDA})~\cite{blei2003latent} and Document Embedding (e.g., \textbf{Doc2Vec}~\cite{le2014distributed}), which both map a document from a high-dimensional sparse vector representation (e.g., 1-hot representations) to a low dimensional latent vector representation. Compared to general dimension reduction methods such as PCA, text-based dimension reduction methods may have advantages as they capture important concepts and semantic relations in a document.


\subsection{Matching}
Matching plays an important role in reducing the bias due to confounding variables. 
We investigate different matching techniques: (1) with ground truth matching pairs (e.g., ``identical twins'') (2) matching based on surface features using Nearest Neighbor Matching (NNM) and (3) our proposed approach, matching based on latent features using NNM.  
For a given binary treatment variable $X_T$ (e.g., whether a word appears in a document or not), we assign all the samples satisfying $X_T=1$ to the treated group (e.g., those with a particular word) and those with $X_T=0$ to the control group (those without a particular word).  If $X_T$ is continuous (e.g., the TF*IDF score of a word), we first  transform it to a binary variable (e.g., $\hat{X_T}=1$ if the TF*IDF score is non-zero, or above the mean, and $\hat{X_T}=0$ otherwise). 

We then pair each sample in the treated group with a  sample from the control group.  
Assuming ground truth matching pairs are not available, 
we employ NNM to find the best matching pairs based on a chosen similarity measure (e.g., cosine similarity). For example, in PSM, given the propensity score of a sample from the treated group, we use NNM to identify a sample from the control group whose propensity score is the closest. 
We may also use NNM to match treated and control samples based on latent or surface features. 
In both cases we use cosine similarity due to its success in matching text for information retrieval. 
To improve matching speed when the number of samples or the number of features is large, we employ a KD-tree based search algorithm ~\cite{bentley1975multidimensional} to quickly locate the nearest neighbor. 
 
\subsection{Feature Selection}
After the samples are paired, we need to decide whether there is any significant difference between the treated and control group w.r.t. the outcome. 
Following~\cite{paul2017feature}, if the outcome variable is binary, we employ McNemar's test~\cite{mcnemar1947note}, which has its test statistic as
\begin{equation}
    \chi^2 = \frac{(TN-CP)^2}{TN+CP} \mbox{ ,}
\end{equation}
where $TN$ is the number of samples in the treated group with a negative class label and $CP$ is the number of samples in the control group with a positive class label. 
Based on the $p$-value associated with the McNemar's test, we can select the target feature $X_T$ if the $p$-value is less than $\alpha$, where $\alpha$ is a hyperparameter that decides the threshold for the significance test. If the outcome variable is numerical, we can instead compare the outcome means between the treated and control group using a $t$-test. 

\subsection{Classification}
On the real datasets, we experimented with several commonly used machine learning classifiers: logistic regression (\textbf{LR}), random forest (\textbf{RF}), support vector machine (\textbf{SVM}), and multi-layer perceptron (\textbf{MLP}).  Since \textbf{RF} performed the best on the real datasets, we reported only the results with \textbf{RF}. Algorithm~\ref{alg:Framwork} summarizes our causal feature selection process.
 
\begin{algorithm}[t]
  \caption{The Causal Feature Selection Algorithm}
  \label{alg:Framwork}
  \begin{algorithmic}[1]
    \Require
      A dataset $S$ with labels $Y$, surface feature vectors $X$; a classification method $F$; size of latent vectors $K$; a dimension reduction method $G$, a nearest neighbor matching method \textit{NMM}, a statistical significance test method \textit{SST},a significance level threshold $\alpha$ and a similarity threshold $\beta$;
    \Ensure
     Selected feature list $f\_lst$; Classification performance $pf$
    \State Initialize $f\_lst$ = [];
    \For{each feature $X_t \in X $}
        \State Initialize a matching pair list $p\_lst$ = []; treated group $T$ = [];  control group $C$ = [];
        \For{each $s_i\in S$}
            \State if $X_{ti}==1$, add $s_i$ to T 
            \State else add $s_i$ to C 
        \EndFor
        \State  $L_K$=$G(\{X-X_t\},K)$ where $\{X-X_t\}$ includes all the surface features in $X$ except $X_t$ and $L_K$ 
        \State represents a K-dimensional latent feature space.
        \For{each sample $t_j \in T$ } 
            \State $c_{tj}$=NNM($t_j$,$C$,$Sim_{LK}$) where $Sim_{LK}$ is a similarity measure defined over K-dimensional
            \State latent feature space $L$
            \State if $Sim_{LK}(t_j, c_{tj})>beta$, add the pair ($t_j, c_{tj}$) in  $p\_lst$;
        \EndFor
        \State Assume $Y_{p\_lst(T)}$ is the aggregated outcomes associated with all the treated samples in $p\_lst$ and
        \State $Y_{p\_lst(C)}$ is the aggregated outcomes associated with all the control  samples in $p\_lst$
        \State Calculate the $p$-value using \textit{SST} to assess the difference between $Y_{p\_lst(T)}$ and $Y_{p\_lst(C)}$;
        \State if $p$-value $\leq$ $\alpha$, add  $X_t$ in $f\_lst$;  
    \EndFor
    \State  $\hat{Y}=F(f\_lst)$;
    \State Compute classification performance $pf$ based on $Y$ and $\hat{Y}$
  \end{algorithmic}
\end{algorithm}
\section{Datasets}
To evaluate the effectiveness of our methods, we used two synthetic datasets and two real datasets.

\textbf{Synthetic Datasets:}
Our proposed framework is based on two hypotheses: 
\emph{(1)} matching quality plays a significant role in causal feature selection, and \emph{(2)} dimension reduction is effective in matching treated-control pairs.  To test these hypotheses directly, we designed two synthetic datasets. 

The \textbf{Latent} dataset is designed to test the system performance when we are able to observe the ``latent'' features directly. Since there is no need to employ dimension reduction to uncover latent variables in this setting, we can focus on investigating the impact of matching quality on causal feature selection. 
Specifically, we create 250 samples with 50 ``latent'' features $L$ generated from multi-normal distributions with 0 means. To create ground truth treated-control pairs, we generate another 250 samples by adding small noise to the existing samples. We also create a new binary treatment feature $X_T$ to indicate that the first 250 samples belong to the treated group and the second 250 samples belong to the control group. 
To test the impact of feature selection on classification, we generate a binary outcome variable $Y$ based on $L$ and $X_T$ using a  logistic regression function. We also generate 10 random variables called $X_O$ that are not related to $Y$. 
We repeat the process 50 times to generate 50 slightly different datasets.  


The \textbf{Surface} dataset simulates a more realistic setting where latent features are not observable. 
First we use linear regression functions to generate 100 observable features $X_C$ from the 50 latent features $L$ in the \textbf{Latent} dataset. The outcome variable $Y$ is generated from the treatment feature $X_T$ and the 100 observable features $X_C$ using a logistic function. To simulate a real text dataset, we also generated a large number of normally distributed random variables $X_O$ (a total of 2900 of them) that are not relevant to the outcome variable $Y$. We randomly generated 50 slightly different surface datasets in our study.

\textbf{Real Datasets:}
\noindent The \textbf{movie dataset (Movie)}  contains reviews from IMDB \cite{maas2011learning}. Movies are rated on a 1-10 scale and the reviews with a rating $\geq$ 7 are labeled as positive while reviews rated $\leq$ 4 are labeled as negative. Similarly to~\cite{paul2017feature}, we discard neutral reviews. Since feature selection is most useful in low resource scenarios where the training sample size is smaller than the feature size, we only used a small subset of the IMDB dataset, whose size we varied systematically. 
We adjusted the infrequent word filtering threshold to keep the feature size similar to that of the \textbf{Surface} dataset.



The \textbf{State of the Union (SOTU)} dataset contains the annual speech by the presidents of the United States to the congress.\footnote{https://www.presidency.ucsb.edu/documents/presidential-documents-archive-guidebook/annual-messages-congress-the-state-the-union} We use this dataset to study the impact of the speech on a president's post-address approval rate. Each state of the union address is annotated with an ``increase'' or ``decrease'' label based on a president's Gallup Poll approval rate before and right after the SOTU address.\footnote{https://www.presidency.ucsb.edu/statistics/data/presidential-job-approval} 
We also filtered infrequent words and kept the feature size similar to that of the \textbf{Surface} dataset. 
Table~\ref{samplesummary} summarizes the statistics of these datasets including sample and feature size. 

\begin{table}[t]
\small
\centering
\begin{tabular}{lcc}
\toprule
& \#samples  & \#features \\
\midrule
\textit{Latent} & 500  & 61 \\
\textit{Surface} & 500  & 3,001 \\
\textit{Movie} & 400-1,000  & 3,028 \\
\textit{SOTU} & 54  & 2,969\\
\bottomrule
\end{tabular}
\caption{Summary of the Synthetic and Real Datasets}
\label{samplesummary}
\end{table}

\section{Experiments on Synthetic Datasets}
We use the \textbf{Latent} dataset to test systems when there is no need for dimension reduction. First, we study the performance of Causal Feature Selection (CFS) if it is given ``perfect matches (PM)'' (\textbf{CFS-PM}). 
We compare it with a system that performs CFS  based on a randomly generated matching pairs (\textbf{CFS-RM}) or pairs discovered by NNM using latent features (\textbf{CFS-LM}). We use the \textbf{Surface} dataset to test the impact of various dimension reduction methods such as \textbf{nPCA}, \textbf{sPCA}, \textbf{GRP} and \textbf{MBDL}. \\    
\vspace{-0.01in}
\textbf{Evaluation Metric:}
An evaluation metric based on feature selection decisions (i.e., whether statistical significance $\alpha$ is reached) would be heavily dependent on $\alpha$ and the sample size. 
We circumvent this issue by instead using a rank-based approach, where we rank the features according to their $p$-values. 
Since $X_T$ has a causal relationship with $Y$ and the variables in $X_O$ do not, we compare the ranks of 
$X_T$ and a randomly selected irrelevant feature from $X_O$. If a system typically ranks $X_T$ higher than $X_O$, it is good at identifying causal features. 
Our evaluation measure, called \textbf{Rank Correctness} (\textbf{RC}), is defined as the probability a system correctly ranks $X_T$ higher than  $X_O$ on the 50 randomly generated datasets. \\ 
\vspace{-0.01in}
\textbf{Baselines:} We compare our methods with \textbf{L1} regression, a classic association-based feature selection method, and \textbf{PSM} and \textbf{MDM}, two of the commonly used casual inference methods based on the potential outcomes framework. \textbf{PSM} was also adopted for causal feature selection in ~\cite{paul2017feature}. \\ 
\vspace{-0.01in}
\textbf {Results:} Table \ref{syn1} shows the performance of different models on the \textbf{Latent} dataset. \textbf{CFS-PM}(\textbf{RC}=0.62), \textbf{CFS-LM}(\textbf{RC}=0.62) and \textbf{PSM}(\textbf{RC}=0.57) all work better than \textbf{L1}(\textbf{RC}=0.52, p$<$0.05 based on t-test). \textbf{CFS-PM} and \textbf{CFS-LM} also performed significantly better than \textbf{PSM} (p$<$0.05). Unlike the above causal methods, \textbf{MDM} did not perform well (\textbf{RC}=0.54). Finally, as expected, random matching \textbf{CFS-RM} performed the worst (\textbf{RC}=0.49).
On the \textbf{Surface} dataset (shown in Table~\ref{syn2}),  \textbf{CFS-PM} continued to be the best (\textbf{RC}= 0.78) and \textbf{CFS-RM} the worst (\textbf{RC}=0.44). Among the methods that employ dimension reduction, \textbf{CFS-sPCA50} performed the best (\textbf{RC}=0.68). It significantly outperformed \textbf{CFS-NM} (\textbf{RC}=0.52, p$<$0.01), which matches treated and control samples based on surface features without dimension reduction. It also significantly outperformed all the  baselines: \textbf{PSM} (\textbf{RC}=0.55, p$<$0.01), \textbf{MDM}(\textbf{RC}=0.54, p$<$0.01) and \textbf{L1}(\textbf{RC}=0.51, p$<$0.01). In addition, it performed much better than \textbf{CFS-sPCA10} (\textbf{RC}=0.54, p$<$0.01), which does not have access to the ground truth $K$. Models employing nPCA also performed quite well (e.g., \textbf{CFS-nPCA50}, \textbf{RC}=0.62).

\begin{figure}[t]
\begin{floatrow}

\capbtabbox{%
\scriptsize
\begin{tabular}{lcc}
\cline{1-3}
Methods & \textbf{CFS-PM}  & \textbf{CFS-LM}  \\
\cline{1-3}
\textbf{RC} & 0.62  & 0.62  \\
\cline{1-3}
Methods &\textbf{PSM} &\textbf{CFS-RM}  \\
\cline{1-3}
\textbf{RC} & 0.57 & 0.49  \\
\cline{1-3}
Methods &\textbf{L1} & MDM \\
\cline{1-3}
\textbf{RC} & 0.52 & 0.54 \\
\cline{1-3}
\end{tabular}

}{%
  \caption{Performance on \textit{Latent}}
\label{syn1}
}
\hspace{-4mm}
\capbtabbox{%
\scriptsize
\begin{tabular}{lccccc}
\cline{1-6}
Methods & \textbf{CFS-PM} & \textbf{CFS-NM} & \textbf{L1} &\textbf{PSM} &\textbf{CFS-RM} \\
\cline{1-6}
\textbf{RC} & 0.78  &0.52& 0.51 & 0.55&0.44 \\
\cline{1-6}
\cline{1-6}
Methods & \textbf{CFS-nPCA10}& \textbf{CFS-nPCA50}& \textbf{CFS-sPCA10}& \textbf{CFS-sPCA50} &\textbf{CFS-GRP10} \\
\cline{1-6}
\textbf{RC} & 0.54  & 0.62 & 0.54&0.68& 0.5 \\
\cline{1-6}
\cline{1-6}
Methods& \textbf{CFS-GRP50}  &\textbf{CFS-MBDL10} & \textbf{CFS-MBDL50}& \textbf{MDM}\\
\cline{1-6}
\textbf{RC}   & 0.54 & 0.52&0.56&0.54 \\
\cline{1-6}
\end{tabular}
}{%
  \caption{Performance on \textit{Surface} }
\label{syn2}
}
\end{floatrow}
\end{figure}

In summary, based on the experiments on the synthetic datasets, we found that \emph{(1)} models with access to ground truth ``perfect twins'' performed the best; \emph{(2)} when there is an underlying latent space, dimension reduction can be used to improve model performance over those that match directly based on surface features without dimension reduction; \emph{(3)} our proposed models can significantly outperform all the established baselines including \textbf{L1}, \textbf{MDM} and \textbf{PSM}; and \emph{(4)} models employing the ground truth dimensionality $K$ performed better than those that do not have access to this information. 

\section{Experiments on Real-World Datasets}
Since treatment/control group assignment is a binary decision, on both the \textbf{Movie} and the \textbf{SOTU} dataset we binarize the word count for the treatment variable $X_T$ for feature selection (1 if a word appears in a document and 0 if not). During classification, we use the TF*IDF scores of the selected features as the predicting variables. All the classification results  reported here are based on the random forest model (\textbf{RF}). 
In addition to the models tested on the \textbf{Surface} dataset, we introduced three new models: \textbf{CFS-LDA}, \textbf{CFS-Doc2Vec} and \textbf{Full Model}. Both \textbf{LDA} and \textbf{Doc2Vec} are widely used dimension reduction methods for text analysis.   
The \textbf{Full Model} employs all the TF*IDF word features without any dimension reduction. 
Unlike the synthetic datasets, here we do not know the ground truth on whether a feature is causally related to an outcome or not. Thus, instead of \textbf{RC}, we adopt extrinsic  evaluation measures typically used for text classification: recall, precision, and F1-score.  
 
On the \textbf{Movie} dataset, to investigate how system performance may be impacted by sample size, we systematically varied the sample size from 400 to 1,000. For each dataset, we randomly split the samples into training (80\%) and development (20\%). We used the development data to tune model parameters (e.g., dimensionality $K$ and $\alpha$).  Since the original movie dataset is much larger (with over 50,000 reviews), we randomly selected 100 new reviews from the remaining data as the test dataset. For \textbf{SOTU}, since it is a small dataset, we employ nested cross-validation for hyperparameter tuning and testing. In the outer loop, we employ leave-one-out cross validation to split data into training and testing. In the inner loop, we use five-fold cross validation to split data into training and development. The development data is used for hyperparameter tuning. 
\vspace{-0.01in}
\subsection{Classification Accuracy}
\begin{figure}[t]
\begin{floatrow}
\ffigbox{%
  \includegraphics[width=6cm]{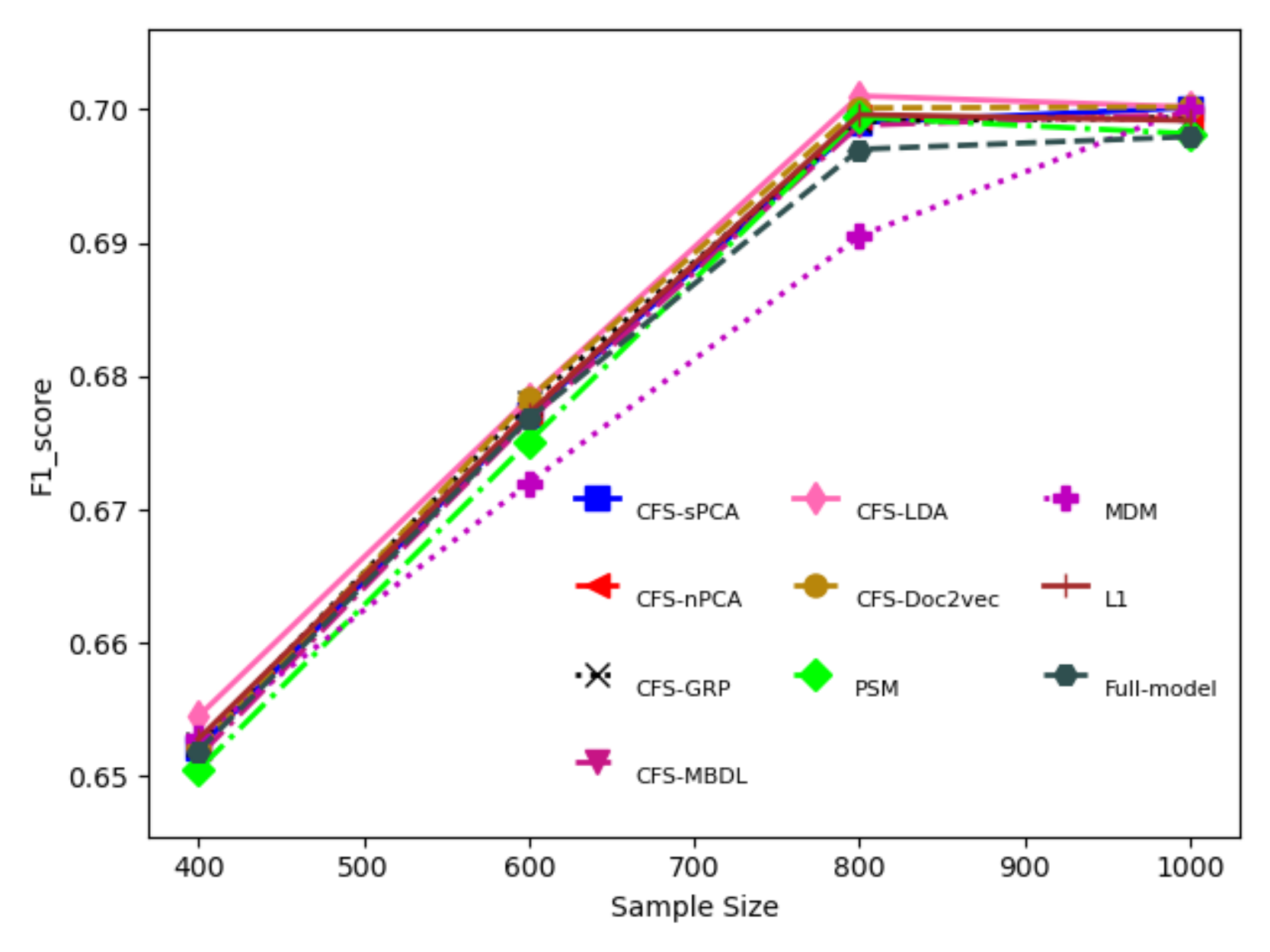}
}{%
  \caption{System Performance (F1) on the \textit{Movie} dataset.}
    \label{movie-f1-fig}
}
\hspace{5mm}
\capbtabbox{%
\small
  \begin{tabular}{lccc}
\cline{1-4}
\diagbox[width=5em,trim=l]{Models}{Metrics}  & Recall & Precision & F1-score\\
\cline{1-4}
\textbf{CFS-nPCA}  & 0.575 & 0.959 & 0.719\\
\textbf{CFS-sPCA} & 0.568 & 0.961 & 0.714\\
\textbf{CFS-GRP}  & 0.570 & 0.941 & 0.710\\
\textbf{CFS-MBDL}  & 0.555 & 0.951 & 0.701\\
\textbf{CFS-LDA}  & \textbf{0.660} & \textbf{0.962} & \textbf{0.783}\\
\textbf{CFS-Doc2vec}  & 0.608 & 0.948 & 0.741\\
\textbf{PSM}  & 0.583 & 0.945 & 0.721\\
\textbf{MDM}  & 0.580 & 0.939 & 0.717\\
\textbf{L1}  & 0.586 & 0.947 & 0.724\\
\textbf{Full-model}  & 0.578 & 0.947 & 0.718\\

\cline{1-4} 
\end{tabular}

}{%
  \caption{Model Performance on the \textit{SOTU} Dataset.}%
  \label{stateofunion-table}
}
\end{floatrow}
\end{figure}

Figure \ref{movie-f1-fig}  shows the classification results on the \textbf{Movie} dataset. When the sample size was small (=400), \textbf{CFS-LDA} 
(F1=0.655) 
performed statistically significantly better than all the baselines, 
i.e. \textbf{L1} (F1=0.652,p$<$0.05), the \textbf{Full model} (F1=0.651, p$<$0.01), \textbf{PSM} (F1=0.651,  p$<$0.001), and \textbf{MDM} (F1=0.653, p$<$0.05). The pattern remained the same when we increased the sample size to 600, and 800. With sample size 600, the F1 for \textbf{CFS-LDA} was 0.678, which is significantly better than \textbf{L1} (F1=0.677, p$<$0.05), the \textbf{Full model} (F1=0.676, p$<$0.01), \textbf{PSM} (F1=0.675, p$<$0.001), and \textbf{MDM} (F1=0.672,  p$<$0.001). With sample size 800, the F1 for \textbf{CFS-LDA} was 0.701, statistically significantly better than \textbf{L1} (F1=0.697, p$<$0.001), the \textbf{Full model} (F1=0.699,p$<$0.05), \textbf{PSM} (F1=0.699,  p$<$0.05), and \textbf{MDM} (F1=0.691, p$<$0.001). Finally, when the sample size was increased to 1000, \textbf{CFS-LDA} and \textbf{CFS-Doc2vec} were the best performing models followed by \textbf{CFS-sPCA} and \textbf{MDM}, although the differences were not statistically significant. In all cases, \textbf{CFS-LDA}  consistently performed the best. 
Table~\ref{stateofunion-table} shows the results on the \textbf{SOTU} dataset. Again, \textbf{CFS-LDA} had the best performance (\textbf{F1}=0.783). It performed statistically significantly better than all the baselines: \textbf{L1} (F1=0.724, p$<$0.05), the \textbf{Full model} (F1=0.718, p$<$0.01), \textbf{PSM} (F1=0.721,  p$<$0.01), and \textbf{MDM} (F1= 0.717, p $<$0.01).
\vspace{-0.01in}
\subsection{Feature Interpretability}
\begin{figure}[!t]
\begin{floatrow}
\capbtabbox{%
\scriptsize
\begin{tabular}{lcccc}
\cline{1-5}
&\textbf{CFS-LDA} & \textbf{PSM} & \textbf{L1} & \textbf{MDM}\\
\cline{1-5}
\multirow{5}{*}{Pos}&great&treasure&think & mass \\
&love&great&hollywood& trust \\
&pleasure&love&way& doubt \\
&enjoy&classic&millionaire& attack \\
&hollywood&wonder&bet & delight\\
\hline
\hline
\multirow{5}{*}{Neg}&inaccuracy&thriller&light & complaint \\
&lose&weak&teacher & tear \\
&issue&monster&fight & weak\\
&complaint&spoil&woman& heard \\
&terrible&technology&pal & realism\\
\cline{1-5} 
\end{tabular}
}{%
  \caption{Top-5 \emph{pos} and \emph{neg} words from the \textit{Movie} data}
\label{p-value-table-1}
}
\hspace{5mm}
\capbtabbox{%
\scriptsize
\begin{tabular}{lcccc}
\cline{1-5}
&\textbf{CFS-LDA} & \textbf{PSM} & \textbf{L1} & \textbf{MDM}\\
\cline{1-5}
\multirow{5}{*}{Pos}& faith&aspiration&  determine & path \\
&god&courage &wheel & inspire\\
&motivation&effort&administer& benjamin \\
&point& fight&undertake& blue\\
&heart&goal&reinforce& burden \\
\hline
\hline
\multirow{5}{*}{Neg}&hitler&air  &fate & congressman\\
&place& cynic &distress & challenge \\
&commit&place &jurisdiction &act \\
&history&poverty  &terror & battle\\
&defeat&walk  & inequality& bone\\
\cline{1-5} 
\end{tabular}

}{%
  \caption{Top-5 \emph{pos} and \emph{neg} words from \textit{SOTU}}
\label{p-value-table-2}
}
\end{floatrow}
\end{figure}

In addition to improving classification performance, causal feature selection can potentially improve interpretability by filtering out spurious associations between words and class labels. 
Tables~\ref{p-value-table-1} and~\ref{p-value-table-2} show the top-5 most significant positive- (\emph{pos}) and negative-coefficient (\emph{neg}) words per method according to $p$-value. Due to the page limit, we only  
report the results for 
\textbf{CFS-LDA}, the best-performing model in our framework, and the three feature selection baselines: \textbf{ L1}, \textbf{PSM} and \textbf{MDM}. On the \textbf{Movie} dataset, except for ``\emph{hollywood},'' almost all the top words selected by \textbf{CFS-LDA} were sentiment words such as \textit{great, love, pleasure, enjoy} and \textit{complaint}. \textbf{PSM}, another causal feature selection method, identified sentiment words such as \textit{great, love} as well as movie domain words such as \textit{classic, thriller, monster, spoil}.  There was no clear pattern in the top words selected by \textbf{L1} and \textbf{MDM}. 
Disturbingly, \textbf{L1} chose \emph{woman} as a \emph{neg} word.  
On \textbf{SOTU}, the \emph{pos} words identified by \textbf{CFS-LDA} contained spiritual and emotional words such as \textit{faith, god, motivation} and \textit{heart}.  The \emph{pos} words identified by \textbf{PSM}  were related to achieving  (e.g., \textit{aspiration, courage, effort, fight, goal}).  The patterns of the words chosen by \textbf{L1} and \textbf{MDM} were unclear. 

\begin{figure*}[t] 
    \centering
    \includegraphics[width=16cm]{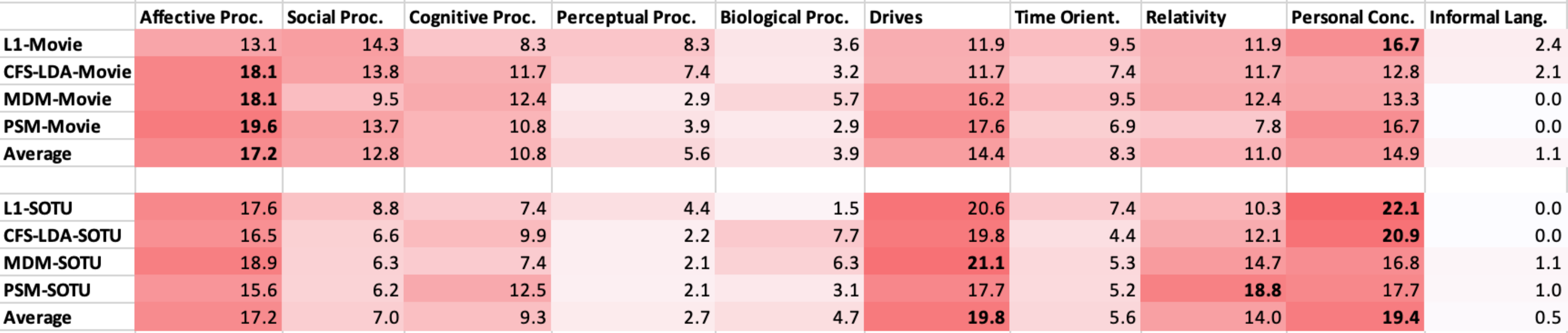}
    \caption{LIWC category percentages for the top 100 words selected by each method. Bold: highest percentages per method. }
    \label{fig:liwc}
\end{figure*}

To gain insight into the types of words selected by each method, we used LIWC~\cite{pennebaker2015development}, a psycholinguistic lexicon, to map the top 100 words 
per method into psychologically and linguistically meaningful categories including  \textit{affective processes}, \textit{cognitive processes},  \textit{drives}, and \textit{personal concerns} (Figure~\ref{fig:liwc}). On the \textbf{Movie} dataset, the causal methods gave the highest attention to \textit{affective processes (18.1\%---19.6\%)}, which we would apriori expect to be the most relevant category for sentiment analysis.  In contrast, the words selected by \textbf{L1} were concentrated on \textit{personal concerns (16.7\%)}, \textit{social processes (14.3\%)}, and with \textit{affective processes (13.1\%)} in third place. 
On the \textbf{SOTU} dataset, except for PSM which had the highest concentration on \textit{relativity}, all the other methods share the same top three LIWC categories: \textit{affective processes}, \textit{drives} and \textit{personal concerns}. Finally, since the methods varied in LIWC percentages, we also computed a stability metric, denoted $sb$, measuring the squared deviation of each method $j$ from the consensus average: 
$sb_j \triangleq mean_{d,c}(mean_{m}(percent_{d,c,m}) - percent_{d,c,j})^2$,  where $j$ and $m$ index methods, $d$ indexes  datasets, and $c$ indexes LIWC categories. Among these methods, \textbf{CFS-LDA} was the most stable, in the sense that its LIWC percentages were closest to the consensus according to the $sb$ metric: \textbf{CFS-LDA}: 1.90, \textbf{MDM}: 2.76,  \textbf{PSM}: 4.19, and \textbf{L1}: 4.54.

\section{Conclusion} 
We have proposed a novel causal feature selection framework which combines dimension reduction with causal inference to identify predictive and interpretable features for text classification.  
Our experiments on both real and synthetic datasets demonstrate the importance of dimension reduction in identifying high-quality matching pairs. The results show that employing commonly used dimension reduction techniques for text data within causal feature selection, e.g. via \textbf{CFS-LDA}, consistently outperforms baselines in all scenarios tested. We have also shown that the word features identified by our methods are easier to interpret versus correlation-based methods. Causal feature selection for interpretable text analysis is still in its infancy. This work is a small but significant step in advancing the state of the art.

\bibliographystyle{unsrtnat}
\bibliography{reference}  






\end{document}